\pdfoutput=1

\documentclass[11pt]{article}

\usepackage{EMNLP2022}

\usepackage{makecell}
\usepackage{times}
\usepackage{latexsym}
\usepackage[norelsize,ruled,linesnumbered]{algorithm2e}
\usepackage{xcolor}

\SetCommentSty{mycommfont}
\usepackage[T1]{fontenc}

\usepackage[utf8]{inputenc}

\usepackage{microtype}
\usepackage{graphicx}
\usepackage{xcolor}
\usepackage{todonotes}
\usepackage{multirow}
\usepackage{amsmath,amssymb}
\usepackage{bbm}
\usepackage{array}

\usepackage{float} 
\usepackage{subfigure}
\usepackage{epstopdf}
\usepackage{tabularx}
\usepackage{inconsolata}

\makeatletter
\renewcommand{\@thesubfigure}{\hskip\subfiglabelskip}
\makeatother

\newcommand{\model}{\textsc{Hegel}}
\newcommand\Tstrut{\rule{0pt}{2.6ex}}         
\newcommand\Bstrut{\rule[-0.9ex]{0pt}{0pt}}   

\title{HEGEL: Hypergraph Transformer for Long Document Summarization}

\author{Haopeng Zhang \and Xiao Liu \and Jiawei Zhang \\       IFM Lab, Department of Computer Science, University of California, Davis, CA, USA \\\texttt{haopeng,xiao,jiawei@ifmlab.org}}

\begin{document}
\maketitle

\begin{abstract}

Extractive summarization for long documents is challenging due to the extended structured input context. The long-distance sentence dependency hinders cross-sentence relations modeling, the critical step of extractive summarization. This paper proposes {\model}, a hypergraph neural network for long document summarization by capturing high-order cross-sentence relations. {\model} updates and learns effective sentence representations with hypergraph transformer layers and fuses different types of sentence dependencies, including latent topics, keywords coreference, and section structure. We validate {\model} by conducting extensive experiments on two benchmark datasets, and experimental results demonstrate the effectiveness and efficiency of {\model}.
\end{abstract}

\section{Introduction}
Extractive summarization aims to generate a shorter version of a document while preserving the most salient information by directly extracting relevant sentences from the original document. With recent advances in neural networks and large pre-trained language models \cite{devlin2018bert,lewis2019bart}, researchers have achieved promising results in news summarization (around 650 words/document) \cite{nallapati2016summarunner,cheng2016neural,see2017get,zhang2022improving,narayan2018ranking,liu2019text}. However, these models struggle when applied to long documents like scientific papers. The input length of a scientific paper can range from $2000$ to $7,000$ words, and the expected summary (abstract) is more than $200$ words compared to $40$ words in news headlines. 

\begin{figure}[!htbp]
\centering
\begin{minipage}[b]{\linewidth}
\includegraphics[width=\textwidth]{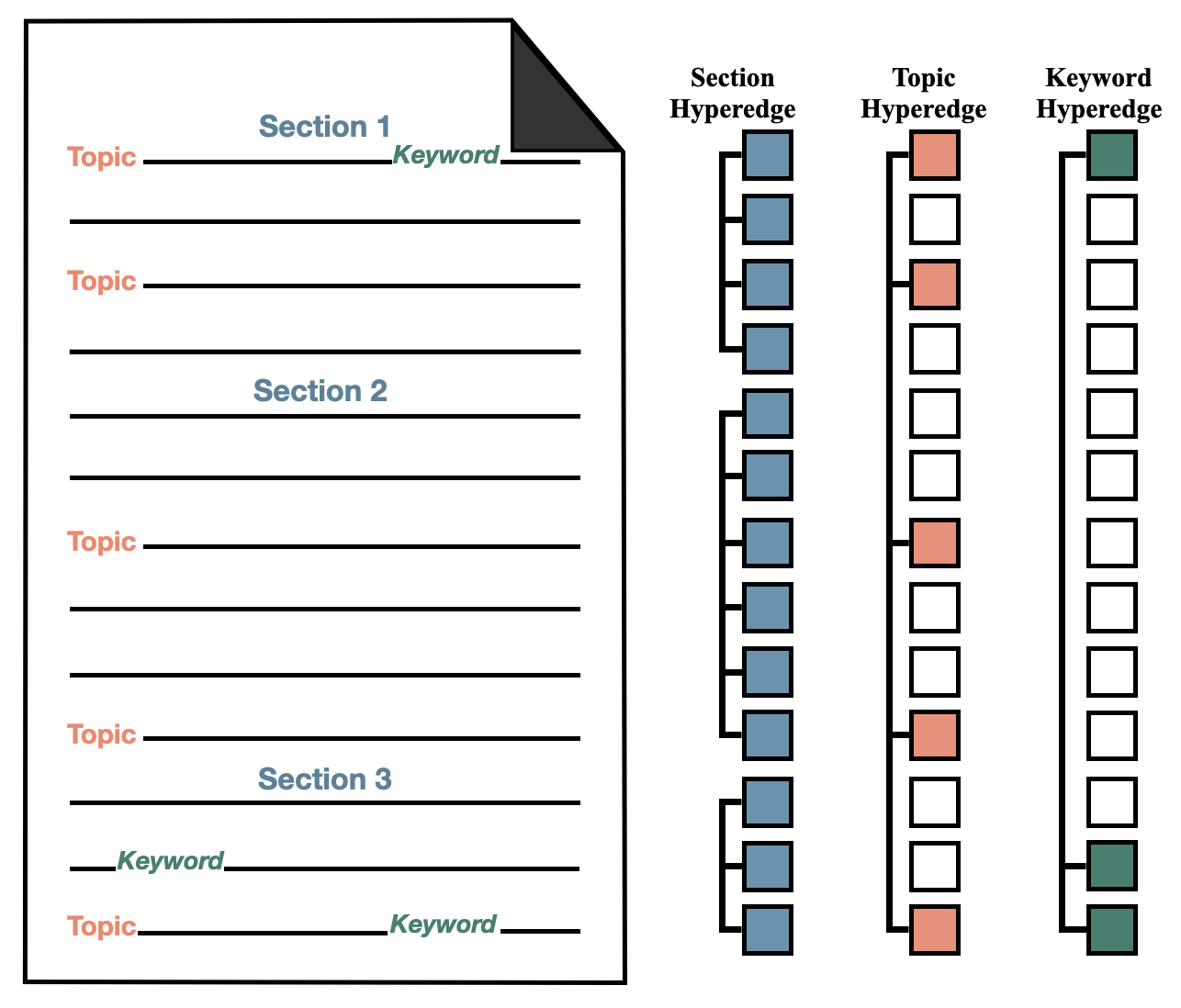} 
\caption{An illustration of modeling cross-sentence relations from section structure, latent topic, and keyword coreference perspectives.}
\label{hyper_edge}
\end{minipage}
\end{figure}

Scientific paper extractive summarization is highly challenging due to the long structured input. The extended context hinders sequential models like RNN from capturing sentence-level long-distance dependency and cross-sentence relations, which are essential for extractive summarization. In addition, the quadratic computation complexity of attention with respect to the input tokens length makes Transformer \cite{vaswani2017attention} based models not applicable. Moreover, long documents typically cover diverse topics and have richer structural information than short news, which is difficult for sequential models to capture.

As a result, researchers have turned to graph neural network (GNN) approaches to model cross-sentence relations. They generally represent a document with a sentence-level graph and turn extractive summarization into a node classification problem. These work construct graph from document in different manners, such as inter-sentence cosine similarity graph in \cite{erkan2004lexrank,dong2020discourse}, Rhetorical Structure Theory (RST) tree relation graph in \cite{xu2019discourse}, approximate discourse graph in \cite{yasunaga2017graph}, topic-sentence graph in \cite{cui-hu-2021-topic-guided} and word-document heterogeneous graph in \cite{wang2020heterogeneous}. However, the usability of these approaches is limited by the following two aspects: \textbf{(1)} These methods only model the pairwise interaction between sentences, while sentence interactions could be triadic, tetradic, or of
a higher-order in natural language \cite{ding2020more}. How to capture high-order cross-sentence relations for extractive summarization is still an open question. \textbf{(2)} These graph-based approaches rely on either semantic or discourses structure cross-sentence relation but are incapable of fusing sentence interactions from different perspectives. Sentences within a document could have various types of interactions, such as embedding similarity, keywords coreference, topical modeling from the semantic perspective, and section or rhetorical structure from the discourse perspective. Capturing multi-type cross-sentence relations could benefit sentence representation learning and sentence salience modeling. Figure~\ref{hyper_edge} is an illustration showing different types of sentence interactions provide different connectivity for document graph construction, which covers both local and global context information.

To address the above issues, we propose {\model} (\textbf{H}yp\textbf{E}r\textbf{G}raph transformer for \textbf{E}xtractive \textbf{L}ong document summarization), a graph-based model designed for summarizing long documents with rich discourse information. To better model high-order cross-sentence relations, we represent a document as a hypergraph, a generalization of graph structure, in which an edge can join any number of vertices. We then introduce three types of hyperedges that model sentence relations from different perspectives, including section structure, latent topic, and keywords coreference, respectively. We also propose hypergraph transformer layers to update and learn effective sentence embeddings on hypergraphs. We validate {\model} by conducting extensive experiments and analyses on two benchmark datasets, and experimental results demonstrate the effectiveness and efficiency of {\model}. We highlight our contributions as follows:

\textbf{(i)} We propose a hypergraph neural model, {\model}, for long document summarization. To the best of our knowledge, we are the first to model high-order cross-sentence relations with hypergraphs for extractive document summarization.
    
\textbf{(ii)} We propose three types of hyperedges (section, topic, and keyword) that capture sentence dependency from different perspectives. Hypergraph transformer layers are then designed to update and learn effective sentence representations by message passing on the hypergraph.

\textbf{(iii)} We validate {\model} on two benchmarked datasets (arXiv and PubMed), and the experimental results demonstrate its effectiveness over state-of-the-art baselines. We also conduct ablation studies and qualitative analysis to investigate the model performance further.

\section{Related Works}
\subsection{Scientific Paper Summarization}

With the promising progress on short news summarization, research interest in long-form documents like academic papers has arisen. \citet{cohan2018discourse} proposed benchmark datasets ArXiv and PubMed, and employed pointer generator network with hierarchical encoder and discourse-aware decoder. \citet{xiao2019extractive} proposed an encoder-decoder model by incorporating global and local contexts. \citet{ju2021leveraging} introduced an unsupervised extractive approach to summarize long scientific documents based on the Information Bottleneck
principle. \citet{dong2020discourse} came up with an unsupervised ranking model by incorporating hierarchical graph representation and asymmetrical positional cues. Recently, \citet{ruan2022histruct+} proposed to apply pre-trained language model with hierarchical structure information.

\begin{figure*}
\vspace{-10pt}
    \centering
    \subfigure[(a)]{
        \includegraphics[width=0.7\textwidth]{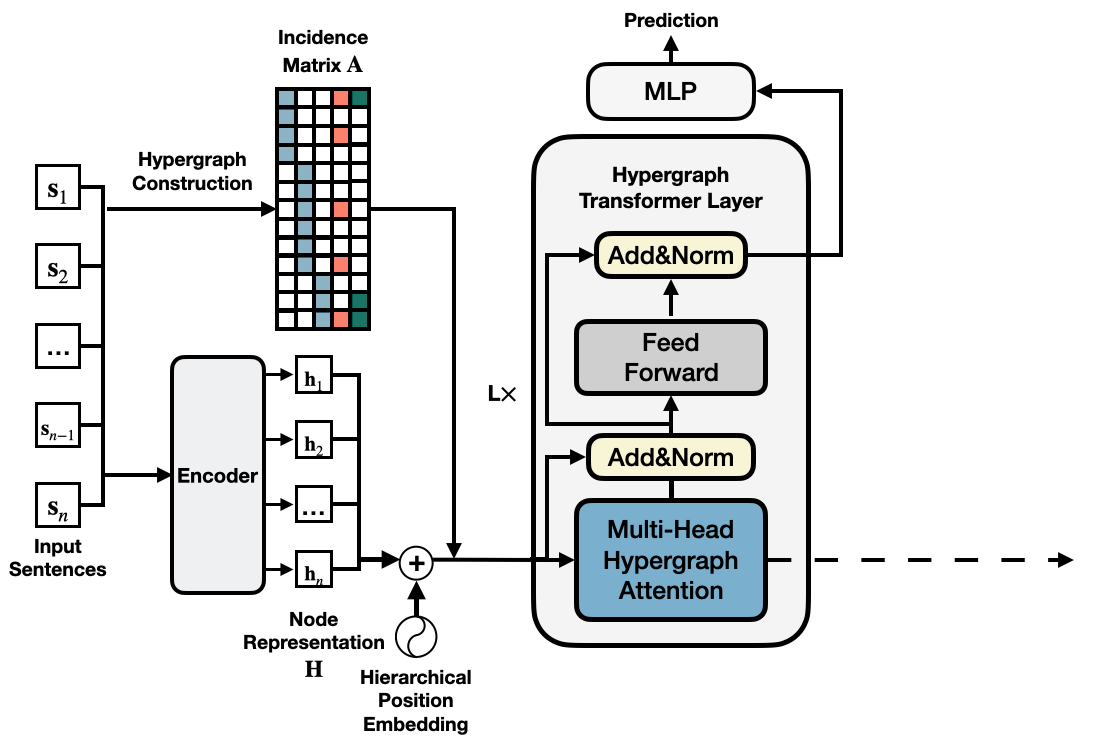}
        \label{model_arch}
    }\hspace{-2mm}
    \subfigure[(b)] {
        \includegraphics[width=0.28\textwidth]{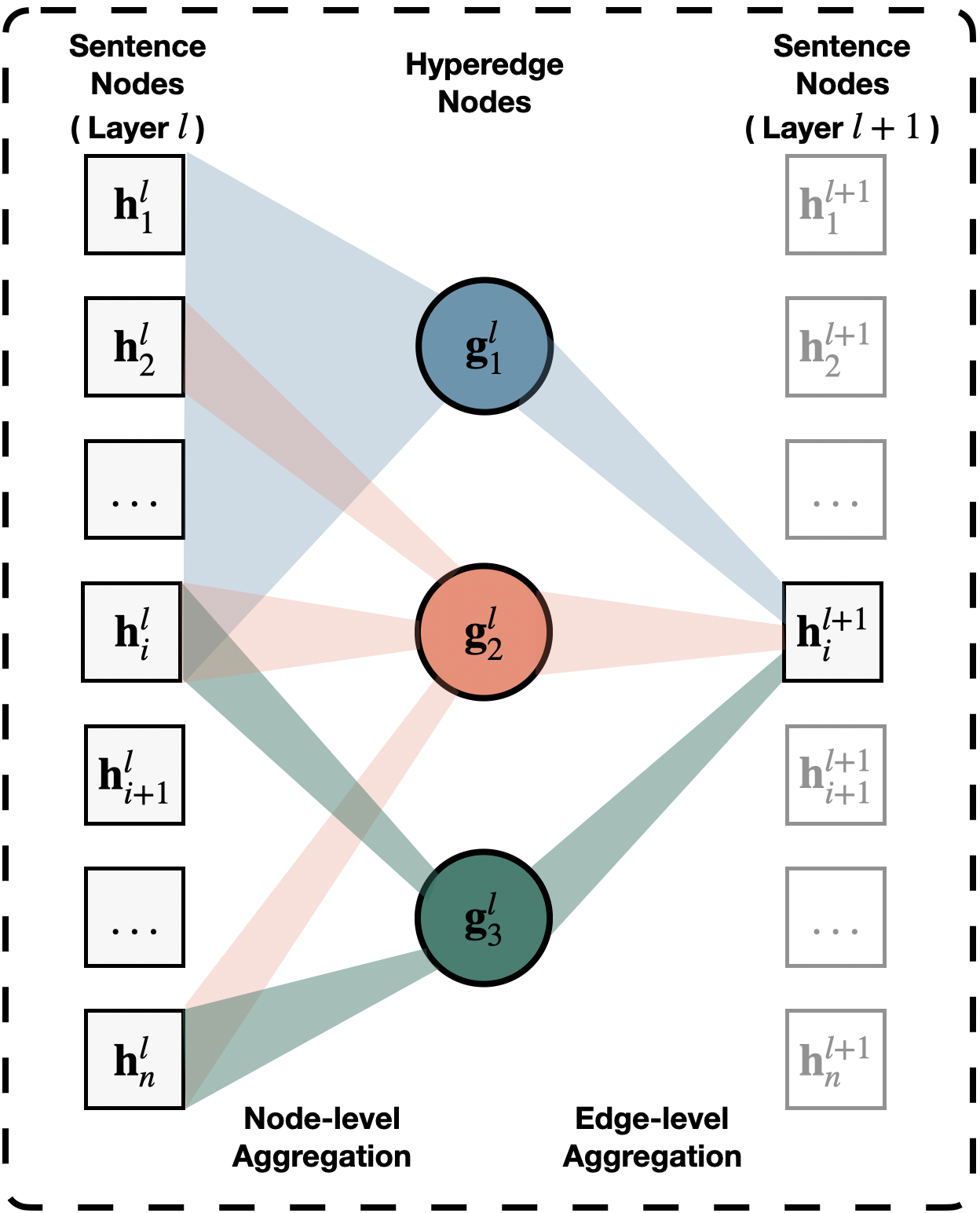}
        \label{layer}
    }
    \caption{(a) The overall  architecture of {\model}. (b) Two-phase message passing in hypergraph transformer layer}
\end{figure*}

\subsection{Graph based summarization}
Graph-based models have been exploited for extractive summarization to capture cross-sentence dependencies. Unsupervised graph summarization methods rely on graph connectivity to score and rank sentences \cite{radev2004centroid,zheng2019sentence,dong2020discourse}. Researchers also explore supervised graph neural networks for summarization. \citet{yasunaga2017graph} applied Graph Convolutional Network (GCN) on the approximate discourse graph. \citet{xu2019discourse} proposed to apply GCN on structural discourse graphs based on RST trees and coreference mentions. \citet{cui2020enhancing} leveraged topical information by building topic-sentence graphs. Recently, \citet{wang2020heterogeneous} proposed to construct word-document heterogeneous graphs and use word nodes as the intermediary between sentences. \citet{jing2021multiplex} proposed to use multiplex graph to consider different sentence relations. Our paper follows this line of work on developing novel graph neural networks for single document extractive summarization. The main difference is that we construct a hypergraph from a document that could capture high-order cross-sentence relations instead of pairwise relations, and fuse different types of sentence dependencies, including section structure, latent topics, and keywords coreference. 
\section{Method}

In this section, we introduce {\model} in great detail. We first present how to construct a hypergraph for a given long document. After encoding sentences into contextualized representations, we extract their section, latent topic, and keyword coreference relations and fuse them into a hypergraph. Then, our hypergraph transformer layer will update and learn sentence representations according to the hypergraph. Finally, {\model} will score the salience of sentences based on the updated sentence representations to determine if the sentence should be included in the summary. The overall architecture of our model is shown in Figure~\ref{model_arch}.

\subsection{Document as a Hypergraph}
A hypergraph is defined as a graph $G=(\mathcal{V}, \mathcal{E})$, where $\mathcal{V}=$ $\left\{v_{1}, \ldots, v_{n}\right\}$ represents the set of nodes, and $\mathcal{E}=\left\{e_{1}, \ldots, e_{m}\right\}$ represents the set of hyperedges in the graph. Here each hyperedge $e$ connects two or more nodes (i.e., $\sigma(e) \geq 2)$. Specifically, we use the notations $v \in e$ and $v \notin e$ to denote node $v$ is connected to hyperedge $e$ or not in the graph $G$, respectively. The topological structure of hypergraph can also be represented by its incidence matrix $\mathbf{A} \in \mathbb{R}^{n \times m}$:

\begin{equation}
\mathbf{A}_{i j}=\left\{\begin{array}{l}
1, \text { if } v_{i} \in e_{j} \\
0, \text { if } v_{i} \notin e_{j}
\end{array}\right.
\end{equation}

Given a document $D = \{s_1, s_2, ..., s_n\}$, each sentence $s_i$ is represented by a corresponding node $v_i \in \mathcal{V}$. A Hyperedge $e_j$ will be created if a subset of nodes $\mathcal{V}_{j}\subset \mathcal{V}$ share common semantic or structural information. 

\subsubsection{Node Representation}
We first adopt sentence-BERT \cite{reimers2019sentence} as sentence encoder to embed the semantic meanings of sentences as $\mathbf{X}=\{\mathbf{x}_1, \mathbf{x}_2, ..., \mathbf{x}_n\}$. Note that the sentence-BERT is only used for initial sentence embedding, but not updated in {\model}.





To preserve the sequential information, we also add positional encoding following Transformer \cite{vaswani2017attention}. We adopt the hierarchical position embedding \cite{ruan2022histruct+}, where position of each sentence $s_i$ can be represented as two parts: the section index of the sentence $p^{sec}_i$, and the sentence index in its corresponding section $p^{sen}_i$.
The hierarchical position embedding (HPE) of sentence $s_i$ can be calculated as:
\begin{equation}
    \text{HPE}(s_i) = \gamma_1 \text{PE}(p^{sec}_i) + \gamma_2 \text{PE}(p^{sen}_i),
\end{equation}
where $\gamma_1, \gamma_2$ are two hyperparameters to adjust the scale of positional encoding and $\text{PE}(\cdot)$ refers to the position encoding function:

\begin{align}
    \text{PE}(pos, 2i) = \sin(pos/10000^{2i/d_{model}}),\\
    \text{PE}(pos, 2i+1) = \cos(pos/10000^{2i/d_{model}}).
\end{align}
Then we can get the initial input node representations $\mathbf{H}^0 = \{\mathbf{h}_1^0, \mathbf{h}_2^0, ..., \mathbf{h}_n^0\}$, with vector $\mathbf{h}_i^0$ defined as:
\begin{equation}
    \mathbf{h}_i^0 = \mathbf{x}_i + \text{HPE}(s_i)
\end{equation}

\subsubsection{Hyperedge Construction}

To effectively model multi-type cross-sentence relations in a long context, we propose the following three hyperedges. These hyperedges could capture high-order context information via the multi-node connection and model both local and global context through document structures from different perspectives. 

\paragraph{Section Hyperedges:}

Scientific papers mostly follow a standard discourse structure describing the problem, methodology, experiments/results, and finally conclusions, so sentences within the same section tend to have the same semantic focus \cite{suppe1998structure}. To capture the \textit{local} sequential context, we build section hyperedges that consider each section as a hyperedge that connects
all the sentences in this section. Section hyperedges could also address the incidence matrix sparsity issue and ensure all nodes of the graph are connected by at least one hyperedge. Assume a document has $q$ sections, section hyperedge $e^{sec}_j$ for the $j$-th section can be represented formally in its corresponding incidence matrix $\mathbf{A}_{sec} \in \mathbb{R}^{n \times q}$ as:
\begin{equation}
    A^{sec}_{i j} =  \left\{\begin{array}{l}
1, \text { if } s_i \in e^{sec}_j \\
0, \text { if } s_i \notin e^{sec}_j
\end{array}\right.
\end{equation}
where $A^{sec}_{i j}$ denotes whether the $i$-th sentence is in the $j$-th section.

\paragraph {Topic Hyperedges:} Topical information has been demonstrated to be effective in capturing important content \cite{cui2020enhancing}. To leverage topical information of the document, we first apply the Latent Dirichlet Allocation (LDA) model \cite{blei2003latent} to extract the latent topic relationships between sentences and then construct the topic hyperedge. In addition, topic hyperedges could address the long-distance dependency problem by capturing \textit{global} topical information of the document. After extracting $p$ topics from LDA, we construct $p$ corresponding topic hyperedges $e^{topic}_j$, represented by the entry $A_{ij}^{topic}$ in the incidence matrix $\mathbf{A}_{topic} \in \mathbb{R}^{n \times p}$ as:

\begin{equation}
    A^{topic}_{i j} =  \left\{\begin{array}{l}
1, \text { if } s_i \in e^{topic}_j \\
0, \text { if } s_i \notin e^{topic}_j
\end{array}\right.
\end{equation}
where $A^{topic}_{i j}$ denotes whether the $i$-th sentence belongs to the $j$-th latent topic.

\paragraph {Keyword Hyperedges:} Previous work finds that keywords compose the main body of the sentence, which are regarded as the indicators
for important sentence selection \cite{wang2013domain,li2020keywords}. Keywords in the original sentence provide significant clues for the main points of the sentence. To utilize keyword information, we first extract keywords for academic papers with KeyBERT \cite{grootendorst2020keybert} and construct keyword hyperedges to link the sentences that contain the same keyword regardless of their sequential distance. Like topic hyperedges, keyword hyperedges also capture \textit{global} context relations and thus, address the long-distance dependency problem. After extracting $k$ keywords for a document, we construct $k$ corresponding keyword hyperedges $e^{kw}_j$, represented in the incidence matrix $\mathbf{A}_{kw} \in \mathbb{R}^{n \times k}$ as:

\begin{equation}
    A^{kw}_{i j} =  \left\{\begin{array}{l}
1, \text { if } s_i \in e^{kw}_j \\
0, \text { if } s_i \notin e^{kw}_j,
\end{array}\right.
\end{equation}
where $s_i \in e^{kw}_j$ means the $i$-th sentence contains the $j$-th keyword.

We finally fuse the three hyperedges by concatenation $\|$ and get the overall incidence matrix $\mathbf{A} \in \mathbb{R}^{n \times m}$ as:
\begin{equation}
    \mathbf{A} =\mathbf{A}_{sec}\|\mathbf{A}_{topic}\|\mathbf{A}_{kw},
\end{equation}
where dimension $m = q+p+k$

The initial input node representations $\mathbf{H}^0 = \{\mathbf{h}_1^0, \mathbf{h}_2^0, ..., \mathbf{h}_n^0\}$ and the overall hyperedge incidence matrix $\mathbf{A}$ will be fed into hypergraph transformer layers to learn effective sentence embeddings.

\subsection{Hypergraph Transformer Layer}

The self-attention mechanism in Transformer \cite{vaswani2017attention} has demonstrated its effectiveness for learning text representation and graph representations \cite{velivckovic2017graph,ying2021transformers,ding2020more,zhang2020text,zhang2020graph}. To model cross-sentence relations and learn effective sentence (node) representations in hypergraphs, we propose the Hypergraph Transformer Layer as in Figure ~\ref{layer}. 
\subsubsection{Hypergraph Attention}
Given node representations $\mathbf{H}^0 = \{\mathbf{h}_{1}^0, \mathbf{h}_{2}^0, ..., \mathbf{h}_{n}^0\}$ and hyperedge incidence matrix $\mathbf{A} \in \mathbb{R}^{n \times m}$, a $l$-layer hypergraph transformer computes hypergraph attention (HGA) and updates node representations $\mathbf{H}$ in an iterative manner as shown in Algorithm~\ref{algorithm}.

Specifically, in each iteration, we first obtain all $m$ hyperedge representations $\{\mathbf{g}_1^l, \mathbf{g}_2^l, ..., \mathbf{g}_m^l\}$ as:


\begin{equation} \label{eq9}
\mathbf{g}_{j}^{l}=\text{LeakyReLU}\left(\sum_{v_{k} \in e_{j}} \alpha_{j k} \mathbf{W}_{h} \mathbf{h}_{k}^{l-1}\right),
\end{equation}
\begin{equation} \label{eq10}
\begin{aligned}
\alpha_{j k} &=\frac{\exp \left(\mathbf{w}_{ah}^{\mathrm{T}} \mathbf{u}_{k}\right)}{\sum_{v_{p} \in e_{j}} \exp \left(\mathbf{w}_{ah}^{\mathrm{T}} \mathbf{u}_{p}\right)}, \\
\mathbf{u}_{k} &=\operatorname{LeakyReLU}\left(\mathbf{W}_{h} \mathbf{h}_{k}^{l-1}\right),
\end{aligned}
\end{equation}
where the superscript $l$ denotes the model layer, matrices $\mathbf{W}_h,\mathbf{w}_{ah}$ are trainable weights and $\alpha_{j k}$ is the attention weight of node $v_k$ in hyperedge $e_j$.

The second step is to update node representations $\mathbf{H}^{l-1}$ based on the updataed hyperedge representations $\{\mathbf{g}_1^l, \mathbf{g}_2^l, ..., \mathbf{g}_m^l\}$ by:

\begin{equation} \label{eq11}
\mathbf{h}_{i}^{l}=\text{LeakyReLU}\left(\sum_{v_{i} \in {e}_{k}} \beta_{i j} \mathbf{W}_{e} \mathbf{g}_{k}^{l}\right),
\end{equation}

\begin{equation} \label{eq12}
\begin{aligned}
\beta_{ki} &=\frac{\exp \left(\mathbf{w}_{ae}^{\mathrm{T}} \mathbf{z}_{k}\right)}{\sum_{v_{i} \in {e}_{q}} \exp \left(\mathbf{w}_{ae}^{\mathrm{T}} \mathbf{z}_{i}\right)}, \\
\mathbf{z}_{k} &=\operatorname{LeakyReLU}\left(\left[\mathbf{W}_{e} \mathbf{g}_{k}^{l} \| \mathbf{W}_{h} \mathbf{h}_{i}^{l-1}\right]\right),
\end{aligned}
\end{equation}
where $\mathbf{h}_{i}^{l}$ is the representation of node $v_i$, $\mathbf{W}_{e}, \mathbf{w}_{ae}$ are trainable weights, and $\beta_{ki}$ is the attention weight of hyperedge $e_k$ that connects node $v_i$. $\|$ here is the concatenation operation.
In this way, information of different granularities and types can be fully exploited
through the hypergraph attention message passing processes.

\paragraph {Multi-Head Hypergraph Attention} As in Transformer, we also extend hypergraph attention (HGA) into multi-head hypergraph attention (MH-HGA) to expand the model’s representation subspaces, represented as:

\begin{equation}
\begin{aligned}
     &\text{MH-HGA}(\mathbf{H}, \mathbf{A}) = \sigma(\mathbf{W}_{O} \|_{i=1}^{h} \text{head}_i), \\
     &\text{head}_i = \text{HGA}_i(\mathbf{H}, \mathbf{A}),
\end{aligned}
\end{equation}
where $\text{HGA}(\cdot)$ denotes hypergraph attention, $\sigma$ is the activation function, $\mathbf{W}_{O}$ is the multi-head weight, and $\|$ denotes concatenation.

\subsubsection{Hypergraph Transformer}
After obtaining the multi-head attention, we also introduce the feed-forward blocks (FFN) with residual connection and layer normalization (LN) like in Transformer. We formally characterize the Hypergraph Transformer layer as below:

\begin{equation}
\begin{aligned}
  \mathbf{H}^{\prime(l)} & = \text{LN}(\text{MH-HGA}(\mathbf{H}^{l-1}, \mathbf{A}) + \mathbf{H}^{l-1})\\
  \mathbf{H}^{l} &= \text{LN}(\text{FFN}(\mathbf{H}^{\prime(l)})+\mathbf{H}^{\prime(l)} \\
\end{aligned}
\end{equation}


\begin{algorithm}
\small
  \SetKwData{Left}{left}\SetKwData{This}{this}\SetKwData{Up}{up}
  \SetKwFunction{Union}{Union}\SetKwFunction{FindCompress}{FindCompress}
  \SetKwInOut{Input}{input}\SetKwInOut{Output}{output}

\Input{node representation $\mathbf{H}^{l-1}\in \mathbb{R}^{n \times d }$,\newline  incidence matrix $\mathbf{A}\in \mathbb{R}^{n \times m }$}
\Output{updated representation $\mathbf{H}^l\in \mathbb{R}^{n \times d }$ }
  \BlankLine
  \For{$head=1,2,...,h$ }{
  \tcp{update hyperedges from nodes}
  \For{$j=1,2,...,m$ }{
    \For{node $v_k \in e_j$}{
    compute attention {$\alpha_{jk}$} with Eq.~\ref{eq10}\;
    update hyperedge representation $\mathbf{g}_j^l$ with Eq.~\ref{eq9}\;
    }
  }
  \tcp{update node representations}
    \For{$i=1,2,...,n$ }{
    \For{hyperedge that $v_i \in e_k$}{
    compute attention $\beta_{ki}$ with Eq.~\ref{eq12};
    update node representation $\mathbf{h}_i^l$ with Eq.~\ref{eq11}\;
    }
  }
  }
    \caption{ $\text{MH-HGA}_{head}(\mathbf{H}, \mathbf{A})$}
  \label{algorithm}
\end{algorithm}
\subsection{Training Objective}
After passing $L$ hypergraph transformer layers, we obtain the final sentence node representations $\mathbf{H}^L = \{\mathbf{h}_1^L, \mathbf{h}_2^L, ..., \mathbf{h}_n^L\}$. We then add a multi-layer perceptron(MLP) followed by a sigmoid activation
function indicating the confidence score for selecting each sentence. Formally, the predicted confidence score ${\hat{y}}_i$ for sentence $s_i$ is: 

\begin{equation}
\begin{aligned}
    {\mathbf{z}_i}  = \text{LeakyReLU}({\mathbf{W}_{p1}}{\mathbf{h}_i^L}), \\
    {\hat{y}}_i = \text{sigmoid}(\mathbf{W}_{p2}\mathbf{z}_i),
\end{aligned}
\end{equation}
where $\mathbf{W}_{p1},\mathbf{W}_{p2}$ are trainable parameters. 

Compared with the sentence ground truth label ${y}_{i}$, we train {\model} in an end-to-end manner and optimize with binary cross-entropy loss as:
\small
\begin{equation}
    \mathcal{L} =- \frac{1}{N\cdot N_d}\sum_{d=1}^{N} \sum_{i=1}^{N_{d}}( y_{i} \log {\hat{y}}_i +(1-y_{i}) \log {(1-\hat{y}}_i)),
\end{equation}
\normalsize
where $N$ denotes the number of training instances in the training set, and $N_{d}$ denotes the number of sentences in the document.


\section{Experiment}



\begin{table}
\small
\centering
\begin{tabular}{c |c |c }

\hline & \text { Arxiv } & \text { PubMed } \\
\hline
\# {train} & 201,427 & 112,291 \\
\# {validation} & 6,431 & 6,402 \\
\# {test} & 6,436 & 6,449 \\
\text {avg. document length} & 4,938  & 3,016  \\
\text {avg. summary length} & 203  & 220  \\
\hline
\end{tabular}
\caption{Statistics of PubMed and Arxiv datasets.}
\label{data_stat}
\end{table}

\begin{table*}[htbp]
\vspace{-12pt}
\centering
\small
\begin{tabular}{c | c  c  c| c  c  c}
    \hline \multirow{2}*{Models} & \multicolumn{3}{c|}{{PubMed}} & \multicolumn{3}{c}{{ArXiv}} \Tstrut\Bstrut\\
    \cline{2-7}
    ~ & {ROUGE-1} & {ROUGE-2} &{ROUGE-L}   & {ROUGE-1} & {ROUGE-2} & {ROUGE-L} \Tstrut\Bstrut\\
    \hline 
        ORACLE & 55.05 & 27.48 & 49.11  & 53.88 &  23.05 & 46.54\Tstrut\\
        LEAD & 35.63 & 12.28 & 25.17 & 33.66 & 8.94 & 22.19 \\
    LexRank (2004) & 39.19  & 13.89 & 34.59 & 33.85  & 10.73 & 28.99 \\
    PACSUM (2019) & 39.79 & 14.00 & 36.09 & 38.57 & 10.93 & 34.33 \\
    HIPORANK (2021) & 43.58 & 17.00 & 39.31 & 39.34 & 12.56 & 34.89 \Bstrut\\

    \hline 
    
    Cheng\&Lapata (2016) & 43.89 & 18.53 & 30.17 & 42.24 & 15.97 & 27.88 \Tstrut \\
    SummaRuNNer (2016) & 43.89 & 18.78 & 30.36 & 42.81 & 16.52 & 28.23 \\
    ExtSum-LG (2019) & 44.85 & 19.70 & 31.43 & 43.62 & 17.36 & 29.14 \\
    SentCLF (2020) & 45.01 & 19.91 & 41.16 & 34.01 & 8.71 & 30.41\\
    SentPTR (2020) & 43.30 & 17.92 & 39.47 & 42.32 & 15.63 & 38.06 \\
    ExtSum-LG + RdLoss (2021) & 45.30 & 20.42 & 40.95 & 44.01 & 17.79 & 39.09\\
    ExtSum-LG + MMR (2021) & 45.39 & 20.37 & 40.99 & 43.87 & 17.50 & 38.97 \\
    HiStruct+ (2022) & 46.59 & 20.39 & 42.11 & 45.22 & 17.67 & \textbf{40.16}\Bstrut\\

    \hline PGN (2017) & 35.86 & 10.22 & 29.69 & 32.06 & 9.04 & 25.16 \Tstrut \\
    DiscourseAware (2018) & 38.93 & 15.37 & 35.21 & 35.80 & 11.05 & 31.80\\
    TLM-I+E (2020) & 42.13 & 16.27 & 39.21  & 41.62 & 14.69 & 38.03\\
    \text { DANCER-LSTM (2020) }  & 44.09 & 17.69 & {40.27} & 41.87 & 15.92 & 37.61\\
\text { DANCER-RUM (2020) }  & 43.98 & 17.65 & 40.25 & 42.70 & 16.54 & 38.44\\
    
    \hline 
    \textbf{HEGEL} (ours) & \textbf{47.13} & \textbf{21.00} & \textbf{42.18} & \textbf{46.41} & \textbf{18.17} & 39.89\Tstrut\Bstrut\\
    \hline
\end{tabular}
\caption{Experimental Results on PubMed and Arxiv datasets.}

\label{exp_result}
\end{table*}

This section presents experimental details on two benchmarked academic paper summarization datasets. We
compare our proposed model with state-of-the-art
baselines and conduct detailed analysis to validate
the effectiveness of {\model}.

\subsection{Experiment Setup}

\paragraph{Datsasets} Scientific papers
are an example of long documents with section discourse structure. Here we validate {\model} on two benchmark scientific paper summarization datasets: ArXiv and PubMed \cite{cohan2018discourse}. PubMed contains academic papers from the biomedical domain, while arXiv contains papers from different scientific domains. We use the original train, validation, and testing splits as in \cite{cohan2018discourse}. The detailed statistics of datasets are shown in Table~\ref{data_stat}.

\paragraph{Compared Baselines}
We perform a systematic comparison with state-of-the-art baseline approaches as follows:

\begin{itemize}
\vspace{-6pt}
\item Unsupervised methods: LEAD that selects the first few sentences as summary; graph-based methods LexRank \cite{erkan2004lexrank}, PACSUM \cite{zheng2019sentence}, and HIPORANK \cite{dong2020discourse}. 
\vspace{-6pt}
\item Neural extractive models: encoder-decoder based model Cheng\&Lapata \cite{cheng2016neural} and SummaRuNNer \cite{nallapati2016summarunner}; local and global context model ExtSum-LG \cite{xiao2019extractive} and its variant RdLoss/MMR \cite{xiao2020systematically}; transformer-based models SentCLF, SentPTR \cite{subramanian2019extractive}, and HiStruct+ \cite{ruan2022histruct+}.
\vspace{-6pt}
\item Neural abstractive models: pointer network PGN \cite{see2017get}, hierarchical attention model DiscourseAware \cite{cohan2018discourse}, transformer-based model TLM-I+E \cite{subramanian2019extractive}, and  divide-and-conquer method DANGER \cite{gidiotis-etal-2020-auth}.

\end{itemize}

\subsection{Implementation Details}
We use pre-trained sentence-BERT \cite{reimers2019sentence} checkpoint \textit{all-mpnet-base-v2} as the encoder for initial sentence representations. The embedding dimension is $768$, and the input layer dimension is $1024$. In our experiment, we stack two layers of hypergraph transformer, and each has $8$ attention heads with a hidden dimension of $128$. The output layer's hidden dimension is set to $4096$. We generate at most $100$ topics for each document and filter out the topic and keyword hyperedges that connect less than $5$ sentence nodes or greater than $25$ sentence nodes. For position encodings, we set the rescale weights $\gamma_1$ and $\gamma_2$ to $0.001$.

The model is optimized with Adam optimizer \cite{loshchilov2017decoupled} with a learning rate of 0.0001 and a dropout rate of 0.3. We train the model on an RTX A6000 GPU for 20 epochs and validate after each epoch using ROUGE-1 F-score to choose checkpoints. Early stopping is employed to select the best model with the patience of 3.

Following the standard-setting, we use ROUGE F-scores \cite{lin2003automatic} for performance evaluation. Specifically, ROUGE-1/2 scores measure summary informativeness, and the ROUGE-L score measures summary fluency. Following prior work \cite{nallapati2016abstractive}, we construct extractive ground truth (ORACLE) by greedily optimizing the ROUGE score on the gold-standard abstracts for extractive summary labeling.

\subsection{Experiment Results}

The performance of {\model} and baseline methods on arXiv and
Pubmed datasets are shown in Table~\ref{exp_result}.
The first block lists the extractive ground truth ORACLE and the unsupervised methods. The second block includes recent extractive summarization models, and the third contains state-of-the-art abstractive methods.

The LEAD method has limited performance on scientific paper summarization compared to its strong performance on short news summarization like CNN/Daily Mail \cite{hermann2015teaching} and New York Times \cite{sandhaus2008new}. The phenomenon indicates that academic paper has less positional bias than news articles, and the ground truth sentence distributes more evenly. For graph-based unsupervised baselines, HIPORANK \cite{dong2020discourse} achieves state-of-the-art performance that could even compete with some supervised methods. This demonstrates the significance of incorporating discourse structural information when modeling cross-sentence relations for long documents. In general, neural extractive methods perform better than abstractive methods due to the extended context. Among extractive baselines, transformer-based methods like SentPTR and HiStruct+ show substantial performance gain, demonstrating the effectiveness of the attention mechanism. HiStruct+ achieves strong performance by injecting inherent hierarchical structures into large pre-trained language models Longformer. In contrast, our model {\model} only relies on hypergraph transformer layers for sentence representation learning and requires no pre-trained knowledge.

As shown in Table~\ref{exp_result}, {\model} outperforms state-of-the-art extractive and abstractive baselines on both datasets. The supreme performance of {\model} shows hypergraphs' capability of modeling high-order cross-sentence relations and the importance of fusing both semantic and structural information. We conduct an extensive ablation study and performance analysis next.
\begin{table}[t]
\centering
\small
\begin{tabular}{c| c c c}
    \hline Model & ROUGE-1 & ROUGE-2 & ROUGE-L \Tstrut\Bstrut\\
    \hline full {\model} & \textbf{47.13} & \textbf{21.00} & \textbf{42.18} \\
     w/o Position & 46.86 & 20.05 & 41.91 \\
     w/o Keyword & 46.92  & 20.71 & 42.03 \\
     w/o Topic & 46.35 & 20.30 & 41.48 \\
     w/o Section & 45.63 & 19.30 & 40.71 \\
    \hline
\end{tabular}
\caption{Ablation study results on PubMed dataset.}
\label{abla}
\end{table}

\section{Analysis}


\subsection{Ablation Study}
We first analyze the influence of different components of {\model}. Table~\ref{abla} shows the experimental results of removing hyperedges and the hierarchical position encoding of {\model} on the PubMed dataset. As shown in the second row, removing the hierarchical position embedding hurts the model performance, which indicates the importance of injecting sequential order information. Regarding hyperedges (row 3-5), we can see that all three types of hyperedges (section, keyword, and topic) help boost the overall model performance. Specifically, the performance drops most when the section hyperedges are removed. The hypergraph becomes sparse and hurts its connectivity. This indicates that the section hyperedges, which contain local context information, play an essential role in the information aggregation process. Note that although we only discuss three types of hyperedges (section, keyword, and topic) in this work, it is easy to extend our model with hyperedges from other perspectives like syntactic for future work.

\subsection{Hyperedge Analysis}

\begin{figure}[htbp]
\centering
\begin{minipage}[t]{0.4\textwidth}
\centering
\includegraphics[width=\textwidth]{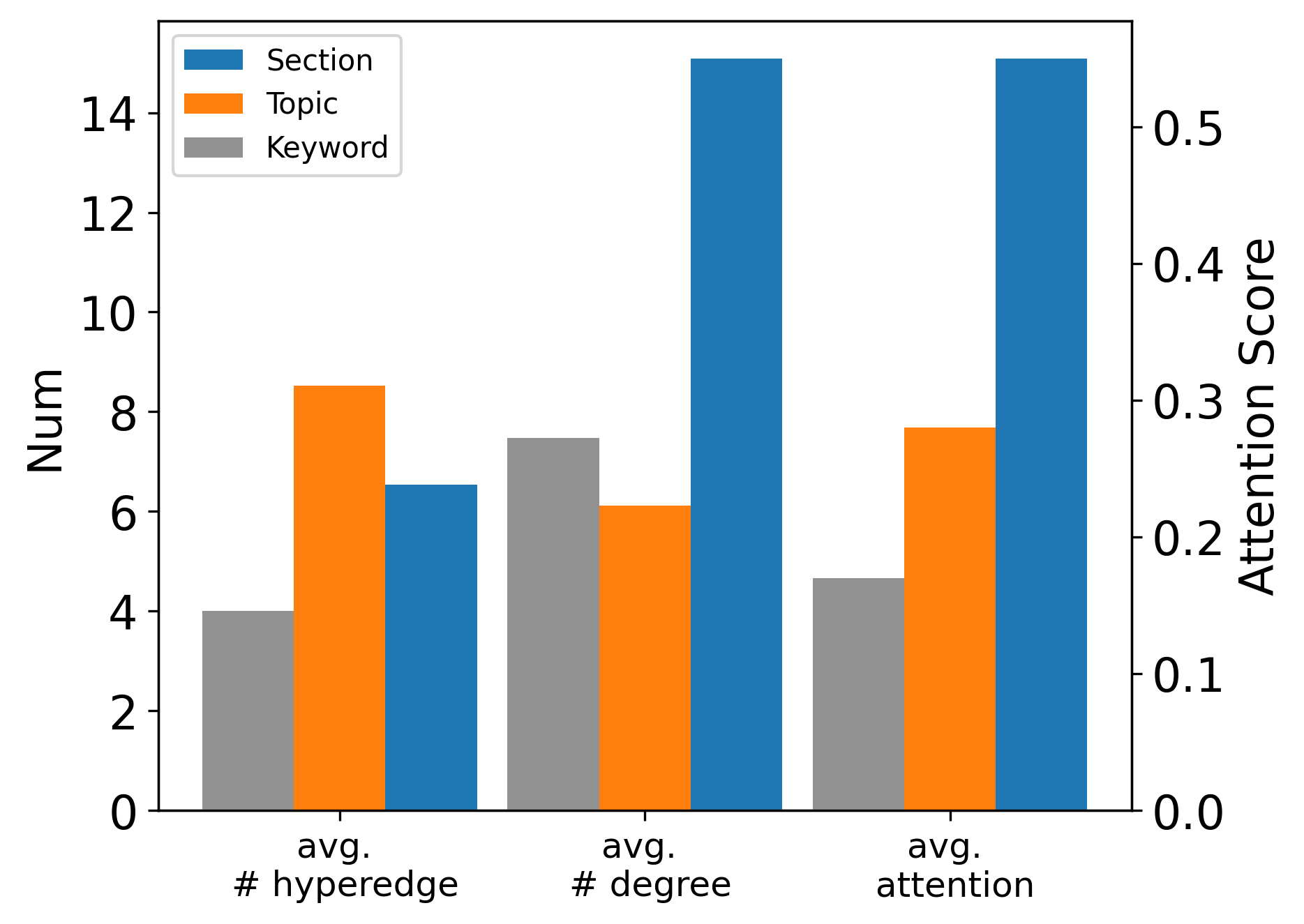}
\caption{Average attention distribution over three types of hyperedges on PubMed dataset.}
\label{fig:attention}
\end{minipage}
\end{figure}

We also explore the hyperedge pattern to understand the performance of {\model} further. As shown in Figure~\ref{fig:attention}, we have the most topic hyperedges on average, and section hyperedges have the largest degree (number of connected nodes). In terms of cross attention over the predicted sentence nodes, {\model} pays more than half of the attention to section hyperedges and pays least to keywords edges. The results are consistent with the earlier ablation study that local section context information plays a more critical role in long document summarization.

\begin{figure}[htbp]
\centering
\begin{minipage}[t]{0.4\textwidth}
\centering
\includegraphics[width=\textwidth]{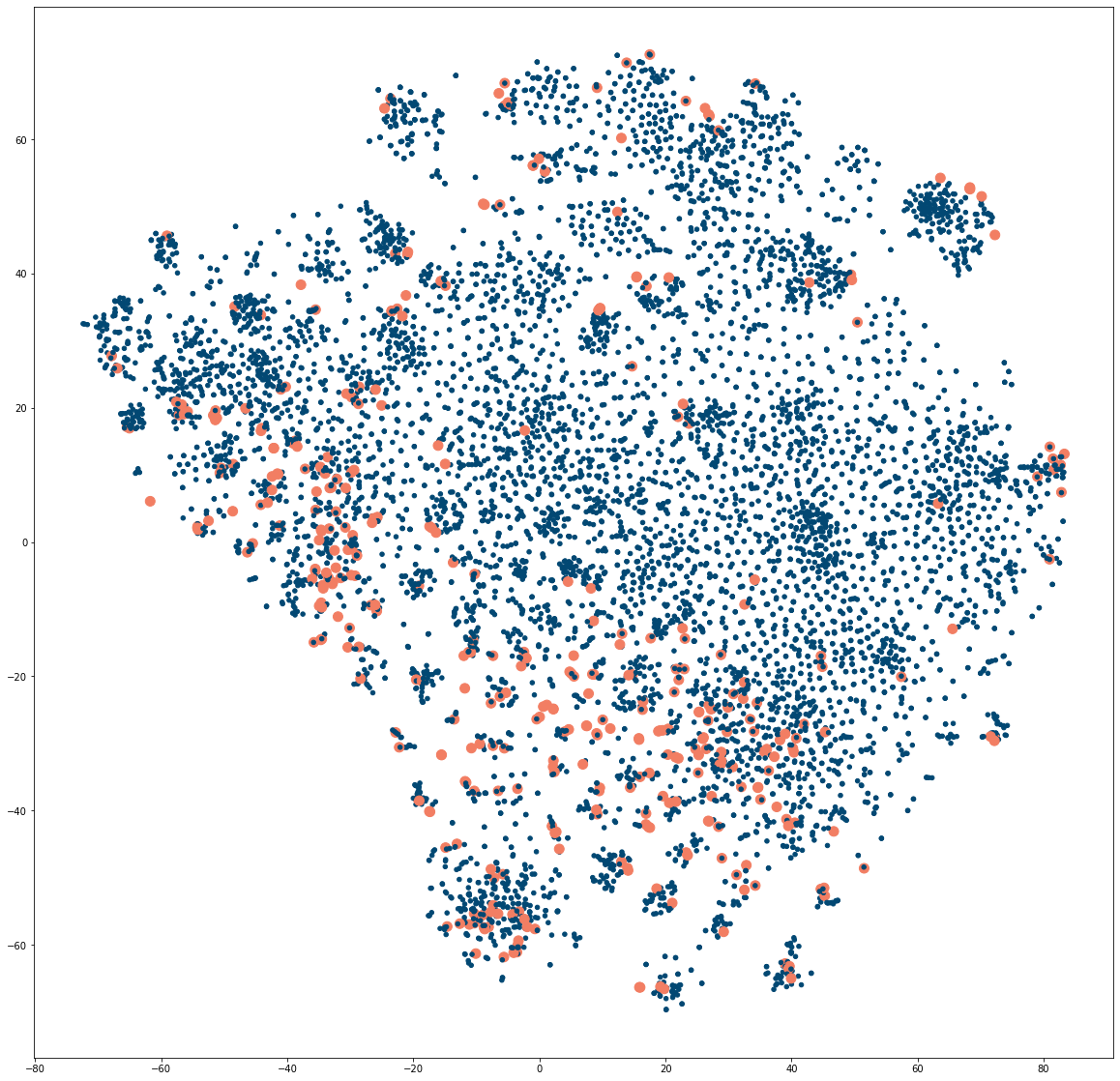}
\caption{Visualization of sentence nodes embeddings for $100$ documents in PubMed test set.}
\label{vis}
\end{minipage}
\end{figure}

\subsection{Embedding Analysis}
To explore the sentence embedding learned by {\model}, we show a visualization of the output sentence node embedding from the last hypergraph transformer layer. We employ T-SNE \cite{JMLR:v9:vandermaaten08a} and reduce each node's dimension to $2$, as shown in Figure~\ref{vis}. The orange dots represent the ground truth sentences, and the blue dots are the non-ground truth sentences. We can see some clustering effects of the ground truth nodes, which also tend to appear in the bottom left zone of the plot. The results indicate that {\model} learns effective sentence embeddings as indicators for salient sentence selection.

\subsection{Case Study}

Here we also provide an example output summary from {\model} in Table~\ref{hyper_att}. We could see that the selected sentences span a long distance in the original document, but are triadically related according to the latent topic and keyword coreference. As a result, {\model} effectively captures high-order cross-sentence relations through multi-type hyperedges and selects these salient sentences according to learned high-order representation.

\begin{table}[htbp]
\centering
\small
\scalebox{1}{
\begin{tabular}{m{0.45\textwidth}}
    \hline\Tstrut\Bstrut\\
     \textcolor[RGB]{241,126,98}{\textbf{[Method]}} Phylogenetic analyses of partial middle east respiratory syndrome coronavirus genomic sequences for \textcolor[RGB]{52,111,96}{\textbf{viruses}} detected in dromedaries imported from oman to united arab emirates, may 2015. (\textcolor[RGB]{108,147,173}{\textbf{Section 1}})\\
     \textcolor[RGB]{241,126,98}{\textbf{[Information]}} Additional information regarding 2 persons with asymptomatic merscov \textcolor[RGB]{52,111,96}{\textbf{infection}} and other persons tested in the study. (\textcolor[RGB]{108,147,173}{\textbf{Section 2}})\\ \textcolor[RGB]{241,126,98}{\textbf{[Information]}} Our findings provide further evidence that asymptomatic human \textcolor[RGB]{52,111,96}{\textbf{infections}} can be caused by zoonotic transmission. (\textcolor[RGB]{108,147,173}{\textbf{Section 2}})\\ \textcolor[RGB]{241,126,98}{\textbf{[Method]}} Merscov genomic sequences determined in this study are similar to those of \textcolor[RGB]{52,111,96}{\textbf{viruses}} detected in 2015 in patients in saudi arabia and south korea with hospital - acquired \textcolor[RGB]{52,111,96}{\textbf{infections}}. (\textcolor[RGB]{108,147,173}{\textbf{Section 3}})\\ \textcolor[RGB]{241,126,98}{\textbf{[Information]}} The \textcolor[RGB]{52,111,96}{\textbf{infected}} dromedaries were imported from oman , which suggests that \textcolor[RGB]{52,111,96}{\textbf{viruses}} from this clade are circulating on the arabian peninsula. (\textcolor[RGB]{108,147,173}{\textbf{Section 4}}) \\
    \hline
\end{tabular}
}
\caption{An example output summary of {\model}. Topics are marked in orange, key words are marked in green, and sections are marked in blue.}
\label{hyper_att}
\end{table}

\section{Conclusion}

This paper presents {\model} for long document summarization. {\model} represents a document as a hypergraph to address the long dependency issue and captures higher-order cross-sentence relations through multi-type hyperedges. The strong performance of {\model} demonstrates the importance of modeling high-order sentence interactions and fusing semantic and structural information for future research in long document extractive summarization.

\section*{Limitations}
Despite the strong performance of {\model}, its design still has the following limitations. First, {\model} relies on existing keyword and topic models to pre-process the document and construct hypergraphs. In addition, we only explore academic paper datasets as a typical example for long document summarization. 

The above limitations may raise concerns about the model's performance. However, {\model} is an end-to-end model, so the pre-process steps do not add the model computation complexity. Indeed, {\model} relies on hyperedge for cross-sentence attention, so it is parameter-efficient and uses $50\%$ less parameters than heterogeneous graph model \cite{wang2020heterogeneous} and $90\%$ less parameters than Longformer-base \cite{beltagy2020longformer}. On the other hand, our experimental design follows a series of previous long document summarization work \cite{xiao2019extractive,xiao2020systematically,subramanian2019extractive, ruan2022histruct+,dong2020discourse,cohan2018discourse} on benchmark datasets ArXiv and PubMed. These two new datasets contain
much longer documents, richer discourse structure than all the news datasets and are therefore ideal test-beds for
long document summarization.



\section*{Acknowledgements}
This work is supported by NSF through grants IIS-1763365 and IIS-2106972. We thank the anonymous reviewers for the helpful feedback.

\bibliography{anthology,custom}
\bibliographystyle{acl_natbib}

\end{document}